\documentclass[conference]{IEEEtrans}
\usepackage{amsmath,amssymb,graphicx,epstopdf,url}

\title{Bayesian Deconvolution of Scanning Electron Microscopy Images Using Point-spread Function Estimation and Non-local Regularization}

\author{
\authorblockN{
Joris Roels\authorrefmark{1}\authorrefmark{2}, 
Jan Aelterman\authorrefmark{1}, 
Jonas De Vylder\authorrefmark{1}, 
Hiep Luong\authorrefmark{1}
Yvan Saeys\authorrefmark{2}\authorrefmark{3}, 
Wilfried Philips\authorrefmark{1}
}
\authorblockA{\authorrefmark{1}Ghent University, Department of Telecommunications and Information Processing, Ghent, Belgium\\\{joris.roels, jan.aelterman, jonas.devylder, hiep.luong, philips\}@telin.ugent.be}
\authorblockA{\authorrefmark{2}VIB, Inflammation Research Center, Ghent, Belgium\\\{yvan.saeys\}@irc.vib-ugent.be}
\authorblockA{\authorrefmark{3}Ghent University, Department of Internal Medicine, Ghent, Belgium}
}

\begin{document}
%\ninept
%
\maketitle
\begin{abstract}
Microscopy is one of the most essential imaging techniques in life sciences. High-quality images are required in order to solve (potentially life-saving) biomedical research problems. Many microscopy techniques do not achieve sufficient resolution for these purposes, being limited by physical diffraction and hardware deficiencies. Electron microscopy addresses optical diffraction by measuring emitted or transmitted electrons instead of photons, yielding nanometer resolution. Despite pushing back the diffraction limit, blur should still be taken into account because of practical hardware imperfections and remaining electron diffraction. Deconvolution algorithms can remove some of the blur in post-processing but they depend on knowledge of the point-spread function (PSF) and should accurately regularize noise. Any errors in the estimated PSF or noise model will reduce their effectiveness. This paper proposes a new procedure to estimate the lateral component of the point spread function of a 3D scanning electron microscope more accurately. We also propose a Bayesian maximum a posteriori deconvolution algorithm with a non-local image prior which employs this PSF estimate and previously developed noise statistics. We demonstrate visual quality improvements and show that applying our method improves the quality of subsequent segmentation steps.
\end{abstract}
\section{Introduction}
\label{sec:intro}
Optical microscopy devices using short wavelength light are physically limited to approximately $200$ nm in lateral and $600$ nm in axial resolution due to photon diffraction. Confocal or super-resolution microscopy significantly alleviates the magnitude of this problem by excitation response or light temporal behavior modeling. Recent developments in this research area have brought optical microscopy into the `nanometer domain' \cite{Mockl2014}. As super-resolution microscopy targets specific biological structures, information of the structural content that was not targeted is mostly not available, limiting the field of view. An alternative to mitigate the diffraction limit without losing a complete sample overview is electron microscopy (EM). Because of the much smaller wavelength of electrons, EM is able to differentiate objects at nanometer scale. The most recent transmission EM (TEM) devices are even capable of imaging at sub-nanometer resolution, allowing researchers to visualize data on a molecular level \cite{Azubel2014}. However, for users interested in volumetric biomedical analysis, 3D scanning EM (SEM) might be a better choice because of its slice by slice imaging workflow, yielding high-resolution 3D images. 

Unfortunately, pushing back the diffraction limit does not allow us to discard blur from the image model. Even under perfect hardware conditions, EM remains physically limited to $1$-$20$ \r{A}ngstr\"om resolution (depending on the accelerating potential) due to electron diffraction, which causes blur at this resolution. Additionally, hardware imperfections such as magnetic lens aberration are nearly unavoidable. The latter source of blur can be reduced by sophisticated expensive lens systems, but even this solution cannot overcome the diffraction limit. An alternative to mitigate blur is post-processing image deconvolution. In this process, the image is modeled as a convolution of the original image and the so-called point spread function (PSF). The original image is then estimated by inverting the convolution (hence `deconvolution'). Two important issues have to be taken into account when applying deconvolution algorithms: sensitivity to PSF estimation errors and improper noise modeling. 

There has been very little research concerning PSF estimation in EM. Most proposed estimators are only valid for specific samples \cite{Hennig2007,Lupini2011}, which makes these techniques inapplicable for general practical purposes. Image deconvolution is a very popular research subject in image restoration including microscopy applications \cite{Lin2013,Lich2013}. These techniques are, in the case of SEM data, more likely to suffer from noise amplification artifacts because of incorrect noise modeling. In contrast, proper image denoising has been actively studied in EM \cite{Sorzano2006,Roels2014}. Nonetheless, deconvolution algorithms exploiting the specific noise and blur characteristics in SEM imaging are hard to find in the literature. In this paper, we propose a PSF estimation procedure for SEM by acquiring images with specific frequency characteristics. Next, we propose a Bayesian deconvolution algorithm with a non-local regularization based on the latter PSF estimation and previously determined noise statistics \cite{Roels2014}, specifically for SEM data. 

In Section \ref{sec:imagemodel}, we will explain our notations and image model. We discuss the estimation of the PSF in Section \ref{sec:psf-em}. Next, we will describe the proposed deconvolution algorithm in Section \ref{sec:deconvolution}. Section \ref{sec:results} illustrates the potential improvements that can be achieved for visualization and image analysis purposes. Lastly, we conclude this paper in Section \ref{sec:conclusion}.

\section{Image model}
\label{sec:imagemodel}
In general, noise should be modeled as a composition of signal-dependent (i.e. Poisson) and signal-independent (Gaussian) noise \cite{Foi2008}. Recent research in SEM has indicated significant noise correlation due to line scanning effects \cite{Roels2014}. Additionally, blur should also be taken into account in the model. We therefore propose a correlation and blur extension of the model proposed in \cite{Foi2008}: 
\begin{equation}\label{eq:image-model}
\mathbf{y} = \mathbf{Hx} + \mathbf{D}(\mathbf{x})\mathbf{C}\mathbf{n}
\end{equation}
where $\mathbf{y}$ and $\mathbf{x}$ represent the acquired and underlying latent image, respectively. The blur kernel (i.e. the PSF of the imaging system) is modeled by a circulant matrix $\mathbf{H}$, $\mathbf{n}$ is a mixed Poisson-Gaussian vector of uncorrelated variables with unit variance, $\mathbf{C}$ is a circulant matrix which models the effects of line scanning noise correlation and $\mathbf{D}(\mathbf{x})$ is a diagonal matrix which models the standard deviation of the noise. According to \cite{Foi2008}, this should be: 
\begin{equation*}
\left(\mathbf{D}(\mathbf{x})\right)_{i,i}= \sqrt{\sigma^2+ \alpha \mathbf{x}_i}. 
\end{equation*}
In this equation, $\sigma^2$ denotes the variance of the signal-independent (Gaussian) part of the noise, $\alpha$ is a positive parameter expressing the `amount' of signal dependency. For specific estimation of the noise related parameters in SEM we refer to \cite{Roels2014}. 

\section{Estimating the SEM PSF}
\label{sec:psf-em}
Before discussing our proposed PSF estimation workflow, it is important to understand how the sample choice can influence PSF estimation accuracy. Therefore, we start with a theoretical discussion on PSF estimation and derive important sample characteristics. 

\subsection{Necessary sample conditions}
\label{sec:sample-necessities}
In principle, the PSF can be estimated by imaging a point-like test sample with known shape and thus known ideal image $\mathbf{x}$. The PSF can then be estimated in the Fourier domain from $\frac{\mathcal{F}(\mathbf{y})(\omega)}{\mathcal{F}(\mathbf{x})(\omega)}$ where $\mathbf{y}$ is the observed image. This approach requires that $\left| \mathcal{F}(\mathbf{x})(\omega) \right|^2$ is non-zero at all frequencies; adequate numeric conditioning also requires that it remains well above zero at all frequencies.
\begin{figure}[t!]
	\centering
	\begin{minipage}{0.49\linewidth}
		\includegraphics[width=\textwidth]{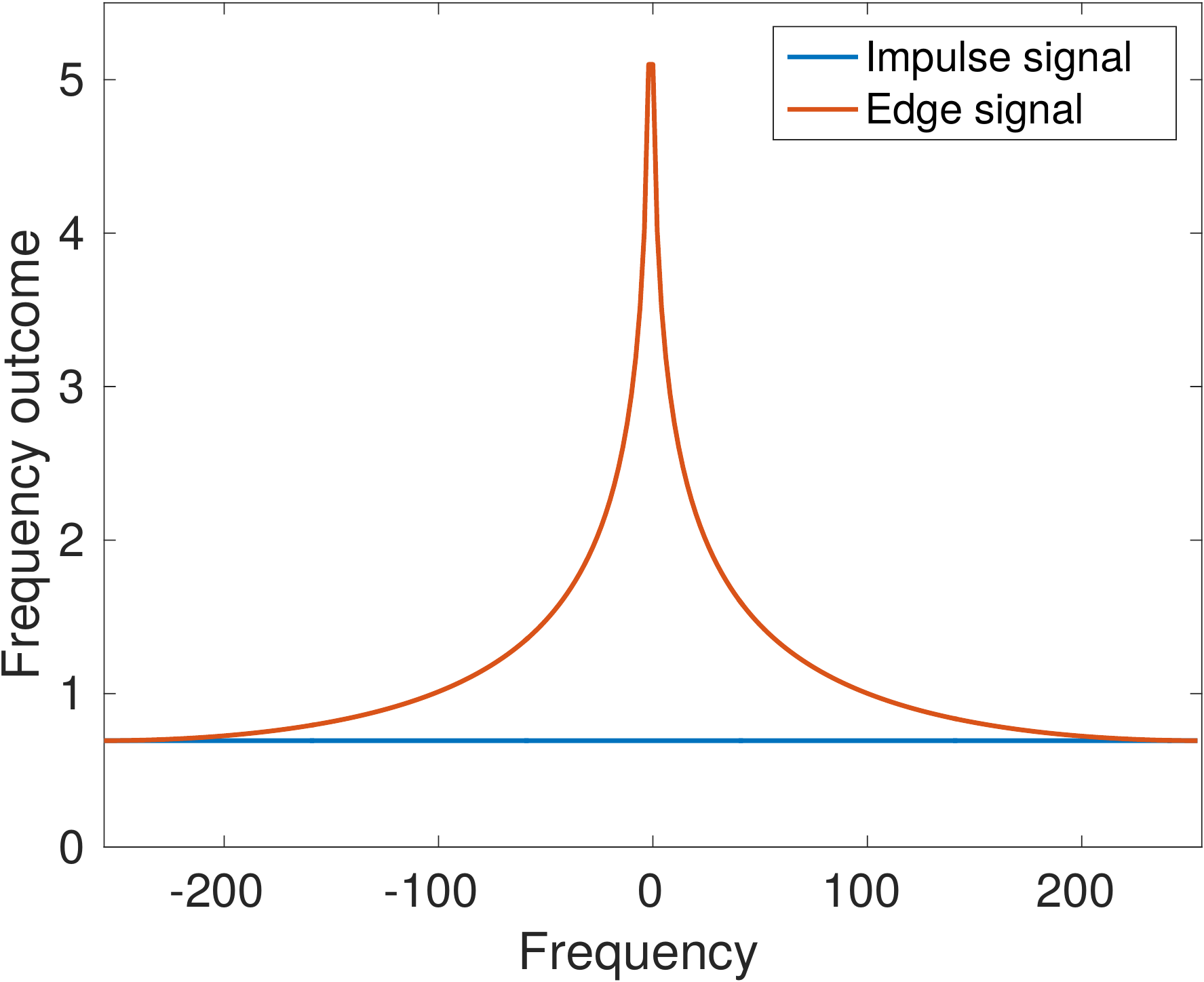}
		\small\centerline{(a)}
			\vspace{0.03cm}
	\end{minipage}
	\begin{minipage}{0.49\linewidth}
		\includegraphics[width=\textwidth]{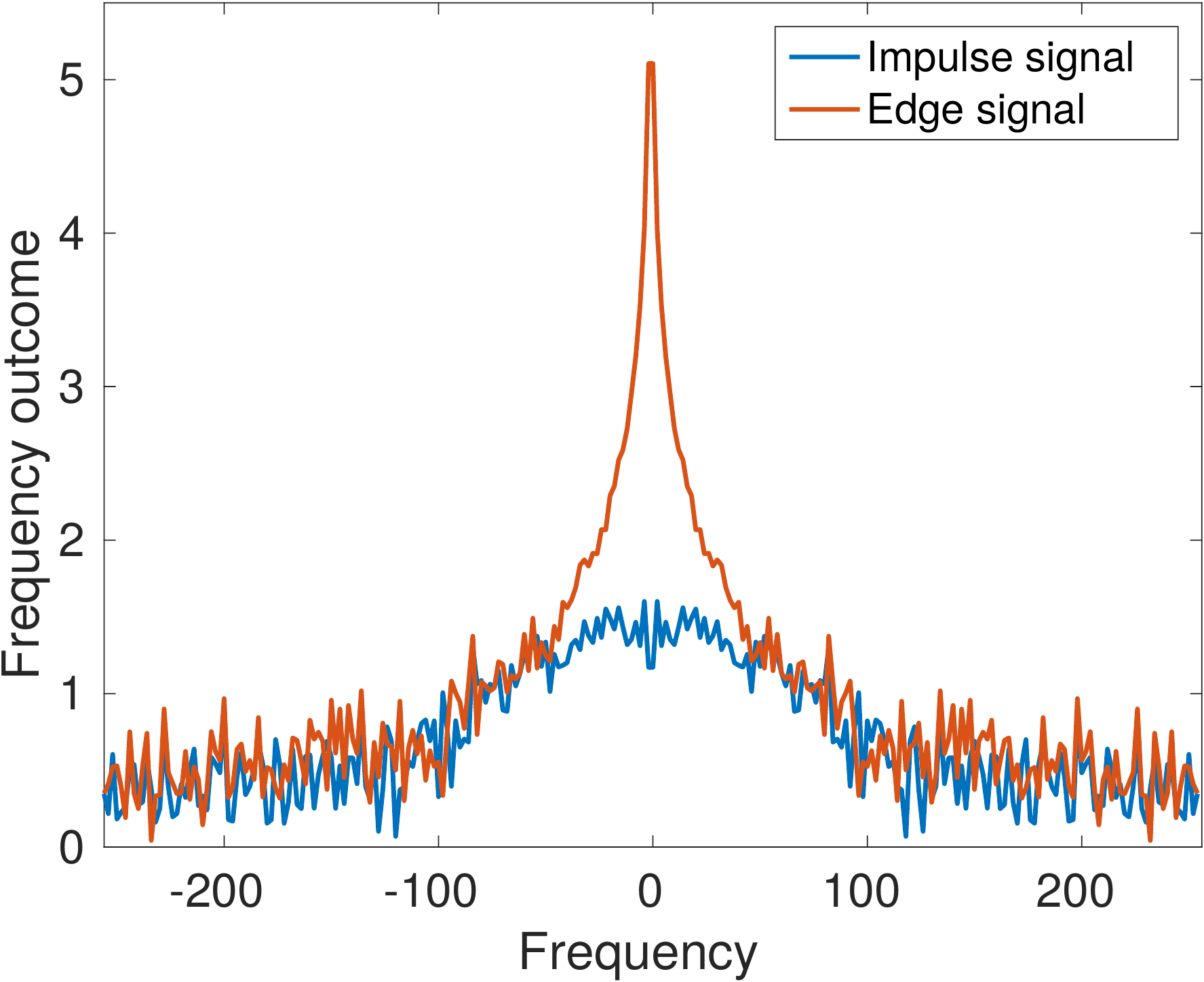}
		\small\centerline{(b)}
			\vspace{0.03cm}
	\end{minipage}
	\begin{minipage}{0.9\linewidth}
		\includegraphics[width=\textwidth]{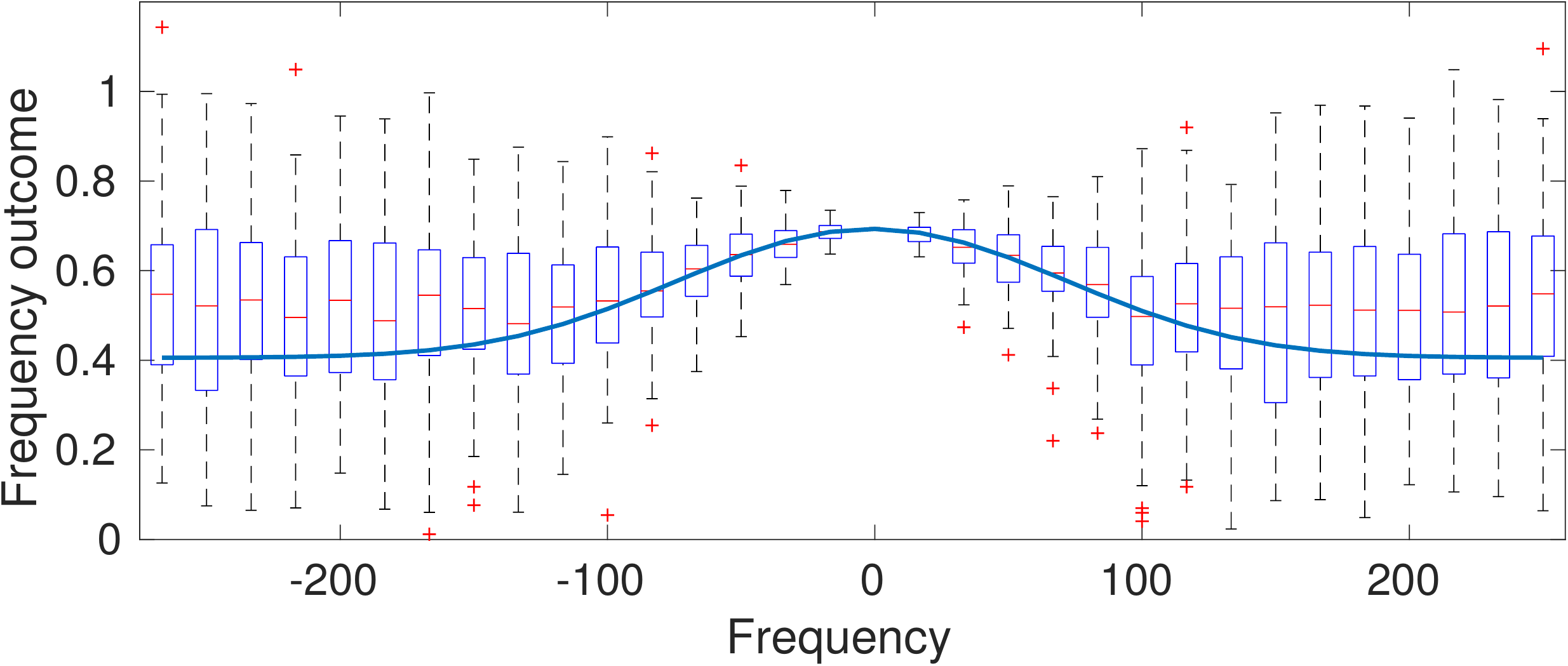}
		\small\centerline{(c)}
	\end{minipage}
	\caption{
		\normalsize
		Frequency spectrum of (a) a one-dimensional impulse and crisp edge signal and (b) noisy, blurred impulse and edge. Figure (c) illustrates the true PSF frequency spectrum $\mathcal{F}(\mathbf{H})(\omega)$ (blue line) and the variability of the corresponding estimated coefficients when random noise is added to the signal. Because of noise, more frequency outcomes approach zero (especially high frequencies), causing significant variability in estimating the PSF coefficients. 
	}
	\label{fig:psf-spectrum}
\end{figure}
Impulse signals are ideal for this purpose because of their corresponding non-zero flat frequency spectrum (Figure \ref{fig:psf-spectrum}a). In practice, a sub-resolution structure of known size and shape is imaged \cite{Gibson1992}. For example, fluorescent beads are commonly used in fluorescence microscopy for this purpose. However, samples like these are less common in EM. Therefore, we acquired SEM images with two expected intensities and predictable geometric characteristics (in our case, crosses, see Figure \ref{fig:PSF-workflow}), which allows us to make a proper estimation of the latent image that should consist of crisp edges switching between the two intensities. These `edge images' also have a complete non-zero spectrum (see Figure \ref{fig:psf-spectrum}a). 

Note however that typically $\mathbf{y}$ also consists of an additional noise signal influencing the PSF estimation as noise may cause small frequency responses for $\mathbf{y}$ (Figure \ref{fig:psf-spectrum}b) that will be greatly amplified in the division with a small frequency response of $\mathbf{x}$ and yield noisy PSF estimations (Figure \ref{fig:psf-spectrum}c). 

To summarize, the acquired images $\mathbf{y}$ should contain as little noise as possible and the underlying latent image $\mathbf{x}$ should have a significant non-zero frequency response for any frequency in order to guarantee a more accurate PSF estimation. 

\subsection{Airy disk}
\label{sec:background}
By examining the physical wave-like behavior of electrons in EM, we expect an Airy disk PSF. This PSF is given by:
\begin{equation}\label{eq:airy-disk}
\mathbf{H}^A(r,\theta) = \frac{1}{Z} \left(\frac{2 J_1 \left(\frac{2 \pi}{\lambda} \text{NA} r\right)}{\left(\frac{2 \pi}{\lambda} \text{NA} r\right)}\right)^2
\end{equation}
where $(r,\theta)$ are polar coordinates relative to the center position of the PSF, $Z$ a normalizing parameter such that $\mathbf{H}^A(r,\theta)$ integrates to $1$, $\lambda$ the electron wavelength, NA the numerical aperture of the EM and $J_1$ the Bessel function of the first kind: 
\begin{equation*}\label{eq:bessel-function}
J_1 (x) = \sum\limits_{m=0}^{\infty} \frac{(-1)^m}{m!\, \Gamma (m+2)} \left( \frac{x}{2} \right)^{2m+1}
\end{equation*}
Note that the Airy disk is rotationally invariant (an important property we will assume in the PSF estimation) and defined for two dimensions; its 1D variant $\mathbf{h}^{A}$ can easily be obtained by evaluating $\mathbf{H}^A(r,\theta)$ for $\theta=0$ and $\theta=\pi$. A second remark is that, according to Equation \ref{eq:airy-disk}, an Airy disk PSF is fixed for a specific EM experiment since the numerical aperture is fixed by the EM and the electron wavelength is determined by the accelerating electrical potential used in the experiment. In practice, because of theoretical conditions that are not always satisfied (imperfect vacuum, relativistic effects, etc.), the observed PSF is a stretched version of Equation \ref{eq:airy-disk}, which can be obtained by the substitution $r \mapsto \tau r$ (where $\tau \in \mathbb{R}^+$). We will denote these PSFs by $\mathbf{H}^{A,\tau}$ and $\mathbf{h}^{A,\tau}$ in the 2D and 1D case, respectively.

\subsection{PSF estimation}
\label{sec:psf-estimation}
\begin{figure}[t!]
\centering
\begin{minipage}{\linewidth}
  \includegraphics[width=\textwidth]{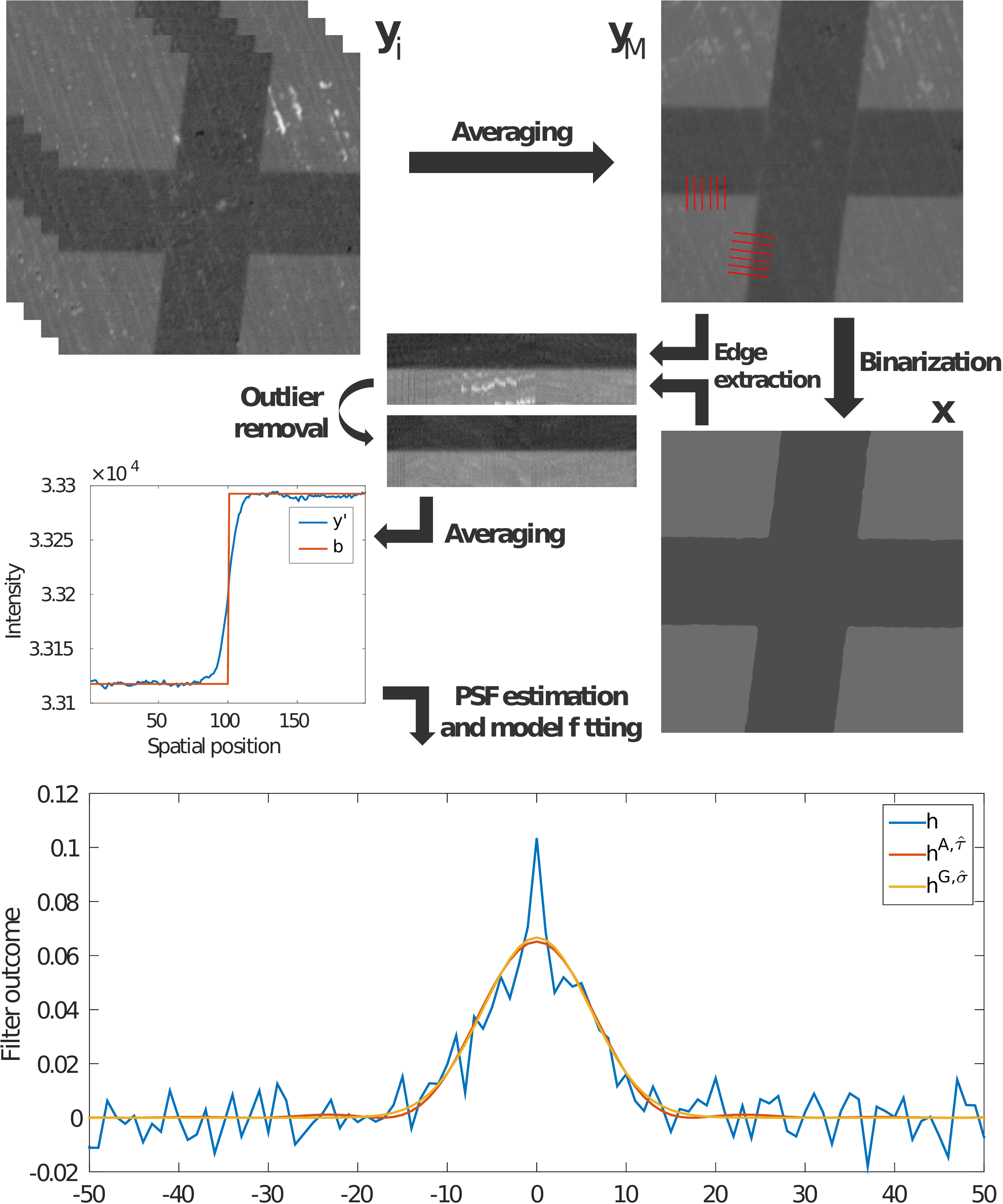}
\end{minipage}
\caption{
\normalsize
PSF estimation workflow
}
\label{fig:PSF-workflow}
\end{figure}

The complete PSF estimation workflow is visualized in Figure \ref{fig:PSF-workflow}. Given an image $\mathbf{y}$ and latent image $\mathbf{x}$ (satisfying the necessities pointed out in Section \ref{sec:sample-necessities} as much as possible), edges are located and combined into a 1D signal. Note, we can restrict the estimation to one dimension because of the assumed rotational invariance of the estimated PSF. The obtained signal is blindly deconvolved and the corresponding PSF estimation is fit to an Airy disk model. Because of the fact that Airy disks are computationally relatively complex, we will also fit the data according to a Gaussian model (denoted by $\mathbf{H}^{G,\sigma}$ and $\mathbf{h}^{G,\sigma}$ in 2D and 1D, respectively), which is a very accurate, yet computationally and mathematically less complex approximation of an Airy disk. 

In our experimental setup, we created cross-like patterns in a homogeneous sample using a focused ion beam. By design, the acquired images should have two intensities (the inside and outside of the crosses) and crisp edges. In practice, however, this will not be the case due to imperfect sample fabrication, noise, blur, etc. The amount of noise is reduced by increasing the dwell-time of the SEM (i.e. the time that is used to detect emitted electrons). Remaining noise is reduced by averaging multiple versions $\mathbf{y}_i$ of the same sample. Note the acquired images require perfect alignment before averaging. This was taken into account by least squares registration in advance. The obtained cross image $\mathbf{y}_M$ contains a smaller amount of noise compared to the original acquired images. The latent image $\mathbf{x}$ is estimated by applying a Gaussian filter to $\mathbf{y}_M$ (in order to remove all the noise) and by Otsu thresholding \cite{Otsu1979} and binarization. 

Given the latent binary image, it is easy to find edges corresponding to a certain direction $\varphi$. Along these edges, we sample one-dimensional signals (red lines in the top right image of Figure \ref{fig:PSF-workflow}) orthogonal to the edge direction up to a certain distance $D$ (in our experiments $D=50$). Note that, because of sample imprecisions, there might be texture in areas that are assumed to have constant intensity. These signals are left out by thresholding the variance in constant intensity areas. If we compute the average of all these signals, we obtain a one-dimensional noise free signal $\mathbf{y}'$ that should be a binary signal  $\mathbf{b}$ (see middle left plot in Figure \ref{fig:PSF-workflow}). The signal is a one-dimensional convolution of the unknown PSF $\mathbf{h}$ and $\mathbf{b}$: 
\begin{eqnarray}\label{eq:convolution}
{\mathbf{y}'}_k &=& \sum\limits_{i=0}^{2D+1} \mathbf{h}_{i} \mathbf{b}_{k-i} \nonumber \\
&=& \mu_1 \sum\limits_{i=0}^{D-k+1} \mathbf{h}_{i} + \mu_2 \sum\limits_{i=D-k}^{2D} \mathbf{h}_{i} 
\end{eqnarray}
where $\mu_1$ and $\mu_2$ are the two unique intensities in $\mathbf{b}$. Note $\mathbf{y}'$ is a vector corresponding to a one-dimensional signal and should not be confused with the vector $\mathbf{y}$ in Equation \ref{eq:image-model}, which corresponds to the two-dimensional acquired image. We remark that Equation \ref{eq:convolution} is a linear system of equations in the unknown variables $\mathbf{h}_{i}$ that can be solved exactly provided $\mu_1 \neq \mu_2$. 

Next, the resulting PSF $\mathbf{h}$ is fit to an Airy disk and Gaussian estimate. These PSFs are completely defined by their stretching parameter $\tau$ and standard deviation $\sigma$, respectively. We minimize the corresponding least square errors: 
\begin{eqnarray*}\label{eq:lse-psf}
\hat{\tau} &=& \arg \min\limits_{\tau} \Vert \mathbf{h}^{A,\tau} - \mathbf{h} \Vert^2 \\
\hat{\sigma} &=& \arg \min\limits_{\sigma} \Vert \mathbf{h}^{G,\sigma} - \mathbf{h} \Vert^2 
\end{eqnarray*}
The fitted PSFs $\mathbf{h}^{A,\hat{\tau}}$ and $\mathbf{h}^{G,\hat{\sigma}}$ are shown at the bottom of Figure \ref{fig:PSF-workflow} and can easily be extended to their corresponding 2D PSF estimates ($\mathbf{H}^{A,\hat{\tau}}$ and $\mathbf{H}^{G,\hat{\sigma}}$, respectively) due to the assumed rotational symmetry. Because of the fact that the Airy disk PSF estimate is very similar to the Gaussian estimate, it is computationally most interesting to use the latter in deconvolution algorithms. 

\section{Proposed deconvolution algorithm}
\label{sec:deconvolution}
Our proposed deconvolution algorithm is based on the non-local means denoising algorithm. We will briefly introduce this technique and its application as an image prior before discussing the proposed deconvolution algorithm. 

\subsection{Non-local means as a Bayesian regularization prior}
\label{sec:nlmeansprior}
The non-local means denoising algorithm (NLMS) \cite{Buades2005} has proven to be very effective in restoring noisy (microscopy) images \cite{Marim2010,Wei2010}. The restored pixel value $\hat{\mathbf{x}}_i$ estimates $\mathbf{x}_i$ as: 
\begin{equation}
\hat{\mathbf{x}}_i = \frac{\sum\limits_{j=0}^{MN-1} w_{i,j} \mathbf{y}_j}{\sum\limits_{j=0}^{MN-1} w_{i,j}} \label{eq:nlms}
\end{equation}
where $\mathbf{y}$ is the observed, noisy image and $w_{i,j}$ are the NLMS weights expressing local similarity between pixels $\mathbf{x}_i$ and $\mathbf{x}_j$. We denote that Equation \ref{eq:nlms} is equivalent to the Bayesian estimator with non-local image prior \cite{Aelterman2012}: 
\begin{equation}
\hat{\mathbf{x}} = \arg\min_{\mathbf{x}}\left\Vert\mathbf{y}-\mathbf{x}\right\Vert_2^2 + \lambda \sum\limits_{i,j=0}^{MN-1} w_{i,j}\left\Vert \left(\mathbf{T}_i -\mathbf{T}_j\right)\mathbf{x} \right\Vert_2^2 \label{eq:nlms-bayesian}
\end{equation}
where $\lambda$ is a regularization parameter and $\mathbf{T}_i \mathbf{x}$ is a vector with $\mathbf{x}_i$ on the first position, i.e. $\left(\mathbf{T}_i \mathbf{x}\right)_j = \delta_{j,0}\mathbf{x}_i$ (where $\delta$ denotes the Kronecker delta). 

In previous work \cite{Roels2014}, we found that noise in SEM is both signal-dependent and highly correlated. We proposed alternative NLMS weights $w_{i,j}'$ in order to take this into account. Therefore, we will use the weights $w_{i,j}'$ proposed in \cite{Roels2014} (NLMS-SC) instead of the original ones that were used in \cite{Buades2005} when deconvolving SEM data.

\subsection{Bayesian deconvolution algorithm}
\label{sec:nlms-scb}
Our deconvolution algorithm is a MAP estimator extending the Bayesian estimator with non-local prior from the previous section to a deconvolution estimator, i.e. the PSF estimate from Section \ref{sec:psf-em} is incorporated as well: 
\begin{equation}\label{eq:deconv-optimization}
\hat{\mathbf{x}} = \arg\min_{\mathbf{x}}\left\Vert\mathbf{y}-\mathbf{Hx}\right\Vert_2^2 + \lambda \sum\limits_{i,j=0}^{MN-1} w_{i,j}'\left\Vert \left(\mathbf{T}_i -\mathbf{T}_j\right)\mathbf{x} \right\Vert_2^2. 
\end{equation}
For fixed $\mathbf{T}_i$ and $\mathbf{T}_j$, the energy function in Equation \ref{eq:deconv-optimization} is a convex function. As a result, an iterative procedure like steepest descent is guaranteed to converge to the minimum in order to solve Equation \ref{eq:deconv-optimization}. Note that this procedure requires an initial solution $\mathbf{x}_{0}$. For this, we used an NLMS-SC filtered version of the acquired image. We denote our algorithm further on with NLMS-SCD.

\section{Results}
\label{sec:results}
Quantitative evaluation of image restoration algorithms on EM data is not straightforward, because of the absence of ground truth images. The main purpose of acquiring EM data is twofold: on the one hand biological researchers want to visualize ultrastructural content as clear as possible, on the other hand the data serves as input for subsequent image analysis, which usually starts with (automated) segmentation. Even though we have convinced biological experts that the images are improved in terms of visual quality, this does not offer a numerical evaluation. Therefore, we also have applied a training based segmentation algorithm on raw and deconvolved 3D SEM data, which can be evaluated quantitatively. 
\begin{figure}[t!]
\centering
\begin{minipage}{0.32\linewidth}
  \includegraphics[width=\textwidth]{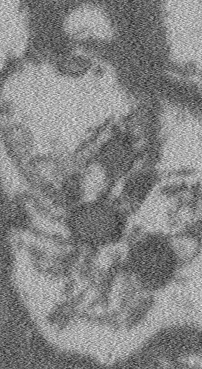}
  \small\centerline{(a)}\label{fig:em_orig_small}
\end{minipage}
\begin{minipage}{0.32\linewidth}
  \includegraphics[width=\textwidth]{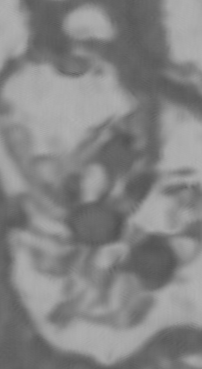}
  \small\centerline{(b)}\label{fig:em_denois_small}
\end{minipage}
\begin{minipage}{0.32\linewidth}
  \includegraphics[width=\textwidth]{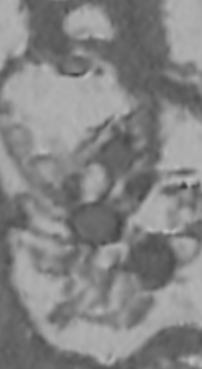}
  \small\centerline{(c)}\label{fig:em_deconv_small}
\end{minipage}
\caption{
\normalsize
Crops of (a) a raw SEM image containing noise and blur, (b) denoised NLMS-SC result and (c) the proposed NLMS-SCD deconvolution. Due to correct PSF modeling, the latter image is considerably sharper than the denoised and original image. 
}
\label{fig:deconv-em}
\end{figure}

\subsection{Visual evaluation}
\label{sec:visual-evaluation}
Figure \ref{fig:deconv-em} shows a noisy and blurred SEM image and the corresponding NLMS-SC denoised and proposed NLMS-SCD deconvolved result. The deconvolved image was obtained using the Gaussian PSF estimate (i.e. $\mathbf{H} = \mathbf{H}^{G,\hat{\sigma}}$) and $\lambda = 0.01$. The NLMS-SC algorithm has removed all the noise, but seems to introduce some edge and texture blurring artifacts. Additionally, blur that was introduced during acquisition has not been suppressed. The NLMS-SCD algorithm solves the latter issue by suppressing noise using non-local means as a regularizer, while jointly modeling the acquired image as a blurred version of the latent image. The resulting NLMS-SCD estimation is considerably sharper than the denoised solution and is revealing ultrastructural details that were previously hard to isolate. 

\subsection{Pre-processing deconvolution for segmentation}
\label{sec:quantitative-evaluation}
Next, we show quantitatively that automated segmentation can be improved by pre-processing deconvolution. The processed data is a $1188 \times 1188 \times 89$ pixel 3D SEM image acquisition of a lung cell. Using the freely available software tool ilastik \cite{Sommer2011}, we train a pixel classifier on the original and deconvolved data. For this, we apply the same training on both data sets (i.e. the same features are computed of the same pixels). This training is solely based on intensity and Laplacian features of a slightly Gaussian (with corresponding standard deviation $\sigma_T$) filtered version of the input data. We then apply the classifier on the complete data sets. As we have manual annotations available of the data, we can evaluate the segmentation quality. For this, we use the recall (R) and Hausdorff distance (HD) metrics. The recall expresses the amount of correctly classified segment pixels compared to the number of incorrectly classified segment pixels. The Hausdorff distance finds the closest segmentation boundary for every ground truth boundary pixel and averages all the corresponding distances. 

The following table shows the average recall and Hausdorff distances along the z-direction. 
\begin{table}[h]
\begin{center}
  \begin{tabular}{| l | c | c | c | c |}
    \hline
        & $\sigma_T$ 	& Raw 				& NLMS-SCD 	 		\\ \hline\hline
    HD 	& 0.7      		& \textbf{1.91}		& 3.02 				\\ \hline
    R 	& 0.7      		& 0.968 			& \textbf{0.979} 	\\ \hline
    HD 	& 1.0      		& \textbf{2.75}		& 3.63 				\\ \hline
    R 	& 1.0 	   		& 0.977 			& \textbf{0.981} 	\\ \hline
    HD 	& 1.6 	   		& \textbf{3.26} 	& 4.01 				\\ \hline
    R 	& 1.6 	   		& 0.980 			& \textbf{0.982} 	\\ \hline
  \end{tabular}
\end{center}
\label{tab:results} 
\end{table}

\begin{figure}[t!]
\centering
\begin{minipage}{0.48\linewidth}
  \includegraphics[width=\textwidth]{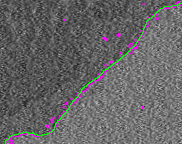}
  \small\centerline{(a)}\label{fig:em_mask_orig}
\end{minipage}
\begin{minipage}{0.48\linewidth}
  \includegraphics[width=\textwidth]{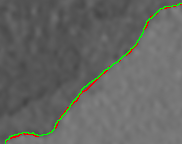}
  \small\centerline{(b)}\label{fig:em_mask_deconv}
\end{minipage}
\caption{
\normalsize
Automatically generated segmentation crop on (a) raw and (b) deconvolved SEM data indicated by purple and red boundary lines, respectively. The ground truth annotations is shown in green. The raw data contains a substantial amount of noise, which leads to irregular edges. 
}
\label{fig:em_masks}
\end{figure}

We denote that the recall is generally higher for the deconvolved images, although the corresponding difference with the original data decreases since features are computed on a more blurry version of the image. This is according to our expectations because the original, noisy data will become very similar to the deconvolved data since they are both being low-pass filtered with increasing $\sigma_T$. Secondly, we notice that the mean Hausdorff distance is smaller for noisy images. This is because the Hausdorff distance searches segment boundaries in the vicinity of the ground truth boundaries. Because of the noise along ground truth edges, the corresponding distances are typically smaller in the raw data (see Figure \ref{fig:em_masks}). For most applications however, a more continuous object border is desired. 

\section{Conclusion}
\label{sec:conclusion}
It has been established that electron microscopy (EM) images typically contain a substantial amount of noise and blur despite the small electron wavelength. Therefore, deconvolution is a crucial step for both visualization and subsequent (automated) image analysis. As deconvolution quality is typically very sensitive to point-spread function (PSF) estimation errors and noise, it is important to model these aspects as accurately as possible. In this paper, we proposed a generic PSF estimation workflow based on the physical expectations of an EM PSF (i.e. an Airy disk) and samples satisfying specific criteria. Secondly, we proposed a Bayesian MAP estimator regularized according to our PSF estimation and previously analyzed noise statistics. We have shown that the restored images can benefit both visualization and image segmentation.

\section*{Acknowledgements}
\label{sec:acknowledgements}
This research has been made possible by the Agency for Flanders Innovation \& Entrepreneurship (VLAIO) and the iMinds ICON BAHAMAS project (\url{http://www.iminds.be/en/projects/2015/03/11/bahamas}). We would like to thank Saskia Lippens (VIB - Bio Imaging Core / Inflammation Research Center) for the provided electron microscopy data and annotations.

%\section{ILLUSTRATIONS, GRAPHS, AND PHOTOGRAPHS}
%\label{sec:illust}
%
%% Below is an example of how to insert images. Delete the ``\vspace'' line,
%% uncomment the preceding line ``\centerline...'' and replace ``imageX.ps''
%% with a suitable PostScript file name.
%% -------------------------------------------------------------------------
%\begin{figure}[htb]
%
%\begin{minipage}[b]{1.0\linewidth}
%  \centering
%  \centerline{\includegraphics[width=8.5cm]{image1.eps}}
%%  \vspace{2.0cm}
%  \centerline{(a) Result 1}\medskip
%\end{minipage}
%%
%\begin{minipage}[b]{.48\linewidth}
%  \centering
%  \centerline{\includegraphics[width=4.0cm]{image2.eps}}
%%  \vspace{1.5cm}
%  \centerline{(b) Results 3}\medskip
%\end{minipage}
%\hfill
%\begin{minipage}[b]{0.48\linewidth}
%  \centering
%  \centerline{\includegraphics[width=4.0cm]{image3.eps}}
%%  \vspace{1.5cm}
%  \centerline{(c) Result 4}\medskip
%\end{minipage}
%%
%\caption{Example of placing a figure with experimental results.}
%\label{fig:res}
%%
%\end{figure}

% \vfill
% \pagebreak

\bibliographystyle{ieeetr}
\bibliography{refs.bib}

\begin{thebibliography}{10}

\bibitem{Mockl2014}
L.~M{\"{o}}ckl, D.~C. Lamb, and C.~Br{\"{a}}uchle, ``{Super-resolved
  Fluorescence Microscopy: Nobel Prize in Chemistry 2014 for Eric Betzig,
  Stefan Hell, and William E. Moerner.},'' {\em Angewandte Chemie
  (International ed. in English)}, pp.~2--8, 2014.

\bibitem{Azubel2014}
M.~Azubel, J.~Koivisto, S.~Malola, D.~Bushnell, G.~L. Hura, A.~L. Koh,
  H.~Tsunoyama, T.~Tsukuda, M.~Pettersson, H.~Hakkinen, and R.~D. Kornberg,
  ``{Electron microscopy of gold nanoparticles at atomic resolution},'' {\em
  Science}, vol.~345, no.~6199, pp.~909--912, 2014.

\bibitem{Hennig2007}
P.~Hennig and W.~Denk, ``{Point-spread functions for backscattered imaging in
  the scanning electron microscope},'' {\em Journal of Applied Physics},
  vol.~102, no.~12, p.~123101, 2007.

\bibitem{Lupini2011}
A.~R. Lupini and N.~de~Jonge, ``{The three-dimensional point spread function of
  aberration-corrected scanning transmission electron microscopy.},'' {\em
  Microscopy and Microanalysis}, vol.~17, no.~5, pp.~817--826, 2011.

\bibitem{Lin2013}
F.~Lin and C.~Jin, ``{An improved Wiener deconvolution filter for
  high-resolution electron microscopy images},'' {\em Micron}, vol.~50,
  pp.~1--6, 2013.

\bibitem{Lich2013}
B.~Lich, X.~Zhuge, P.~Potocek, F.~Boughorbel, and C.~Mathisen, ``{Bringing
  Deconvolution Algorithmic Techniques to the Electron Microscope},'' {\em
  Biophysical Journal}, vol.~104, p.~500a, jan 2013.

\bibitem{Sorzano2006}
C.~Sorzano, E.~Ortiz, M.~L{\'{o}}pez, and J.~Rodrigo, ``{Improved Bayesian
  image denoising based on wavelets with applications to electron
  microscopy},'' {\em Pattern Recognition}, vol.~39, no.~6, pp.~1205--1213,
  2006.

\bibitem{Roels2014}
J.~Roels, J.~Aelterman, J.~{De Vylder}, H.~Q. Luong, S.~Lippens, Y.~Saeys, and
  W.~Philips, ``{Noise analysis and removal in 3D electron microscopy},'' in
  {\em Lecture Notes in Computer Science: Advances in Visual Computing},
  pp.~31--40, 2014.

\bibitem{Foi2008}
A.~Foi, M.~Trimeche, V.~Katkovnik, and K.~Egiazarian, ``{Practical
  poissonian-gaussian noise modeling and fitting for single-image raw-data},''
  {\em IEEE Transactions on Image Processing}, vol.~17, no.~10, pp.~1737--1754,
  2008.

\bibitem{Gibson1992}
S.~F. Gibson and F.~Lanni, ``{Experimental test of an analytical model of
  aberration in an oil-immersion objective lens used in three-dimensional light
  microscopy.},'' {\em Journal of The Optical Society of America A - Optics,
  Image Science and Vision}, vol.~9, no.~1, pp.~154--166, 1992.

\bibitem{Otsu1979}
N.~Otsu, ``{A threshold selection method from gray-level histograms},'' {\em
  IEEE Transactions on Systems, Man, and Cybernetics}, vol.~9, no.~1,
  pp.~62--66, 1979.

\bibitem{Buades2005}
A.~Buades, B.~Coll, and J.-M. Morel, ``{A Non-local Algorithm for Image
  Denoising},'' in {\em Proc. IEEE Computer Society Conference on Computer
  Vision and Pattern Recognition}, vol.~2, pp.~60--65 vol. 2, 2005.

\bibitem{Marim2010}
M.~Marim, E.~Angelini, and J.~C. Olivo-Marin, ``{Denoising in fluorescence
  microscopy using compressed sensing with multiple reconstructions and
  non-local merging},'' in {\em Proc. IEEE Engineering in Medicine and Biology
  Society}, pp.~3394--3397, 2010.

\bibitem{Wei2010}
D.~Y. Wei and C.~C. Yin, ``{An optimized locally adaptive non-local means
  denoising filter for cryo-electron microscopy data},'' {\em Journal of
  Structural Biology}, vol.~172, no.~3, pp.~211--218, 2010.

\bibitem{Aelterman2012}
J.~Aelterman, B.~Goossens, H.~Q. Luong, J.~{De Vylder}, A.~Pizurica, and
  W.~Philips, ``{Combined Non-local and Multi-Resolution Sparsity Prior in
  Image Restoration},'' in {\em Proc. International Conference on Image
  Processing}, pp.~3049--3052, 2012.

\bibitem{Sommer2011}
C.~Sommer, C.~Straehle, U.~K{\"{o}}the, and F.~A. Hamprecht, ``{Ilastik:
  interactive learning and segmentation toolkit},'' in {\em Proc. IEEE
  International Symposium on Biomedical Imaging}, pp.~230--233, 2011.

\end{thebibliography}

\end{document}